\documentclass[sigconf]{acmart}
\AtBeginDocument{%
  \providecommand\BibTeX{{%
    \normalfont B\kern-0.5em{\scshape i\kern-0.25em b}\kern-0.8em\TeX}}}
\usepackage{amsmath}
\usepackage{multicol}
\usepackage{multirow}
\usepackage{algorithm}
\usepackage{algorithmicx}
\usepackage{algpseudocode}

\copyrightyear{2023}
\acmYear{2023}
\setcopyright{rightsretained}
\acmConference[CIKM '23]{Proceedings of the 32nd ACM International Conference on Information and Knowledge Management}{October 21--25, 2023}{Birmingham, United Kingdom}
\acmBooktitle{Proceedings of the 32nd ACM International Conference on Information and Knowledge Management (CIKM '23), October 21--25, 2023, Birmingham, United Kingdom}\acmDOI{10.1145/3583780.3614758}
\acmISBN{979-8-4007-0124-5/23/10}

\begin{document}

\title{A Bipartite Graph is All We Need for Enhancing Emotional Reasoning with Commonsense Knowledge}

\author{Kailai Yang}
\affiliation{%
  \institution{The University of Manchester}
  \city{Manchester}
  \country{United Kingdom}
}
\email{kailai.yang@postgrad.manchester.ac.uk}

\author{Tianlin Zhang}
\affiliation{%
  \institution{The University of Manchester}
  \city{Manchester}
  \country{United Kingdom}
}
\email{tianlin.zhang@postgrad.manchester.ac.uk}

\author{Shaoxiong Ji}
\affiliation{%
  \institution{University of Helsinki}
  \city{Helsinki}
  \country{Finland}
}
\email{shaoxiong.ji@helsinki.fi}

\author{Sophia Ananiadou}
\affiliation{%
  \institution{The University of Manchester}
  \city{Manchester}
  \country{United Kingdom}
}
\email{sophia.ananiadou@manchester.ac.uk}
  
\renewcommand{\shortauthors}{Kailai Yang, Tianlin Zhang, Shaoxiong Ji, and Sophia Ananiadou}

\begin{abstract}
The context-aware emotional reasoning ability of AI systems, especially in conversations, is of vital importance in applications such as online opinion mining from social media and empathetic dialogue systems. Due to the implicit nature of conveying emotions in many scenarios, commonsense knowledge is widely utilized to enrich utterance semantics and enhance conversation modeling. However, most previous knowledge infusion methods perform empirical knowledge filtering and design highly customized architectures for knowledge interaction with the utterances, which can discard useful knowledge aspects and limit their generalizability to different knowledge sources. Based on these observations, we propose a \textbf{B}ipartite \textbf{H}eterogeneous \textbf{G}raph (BHG) method for enhancing emotional reasoning with commonsense knowledge. In BHG, the extracted context-aware utterance representations and knowledge representations are modeled as heterogeneous nodes. Two more knowledge aggregation node types are proposed to perform automatic knowledge filtering and interaction. BHG-based knowledge infusion can be directly generalized to multi-type and multi-grained knowledge sources. In addition, we propose a Multi-dimensional Heterogeneous Graph Transformer (MHGT) to perform graph reasoning, which can retain unchanged feature spaces and unequal dimensions for heterogeneous node types during inference to prevent unnecessary loss of information. Experiments show that BHG-based methods significantly outperform state-of-the-art knowledge infusion methods and show generalized knowledge infusion ability with higher efficiency. Further analysis proves that previous empirical knowledge filtering methods do not guarantee to provide the most useful knowledge information. Our code is available at: \url{https://github.com/SteveKGYang/BHG}.
\end{abstract}





\maketitle

\section{Introduction}

Understanding human emotions is at the core of affective computing. In natural language processing, recent years have witnessed growing research interests in machine's context-aware emotional reasoning ability, especially in conversations, due to its vital importance in scenarios such as empathetic dialogue systems~\citep{ma2020survey} and online opinion mining from social media~\citep{chatterjee2019understanding}. This goal is mostly specified as recognizing the emotion~\citep{DBLP:journals/access/PoriaMMH19} or the emotion cause~\citep{poria2021recognizing} of certain utterances within a conversation.

\begin{figure}[htpb]
\centering
\includegraphics[width=8cm,height=6.4cm]{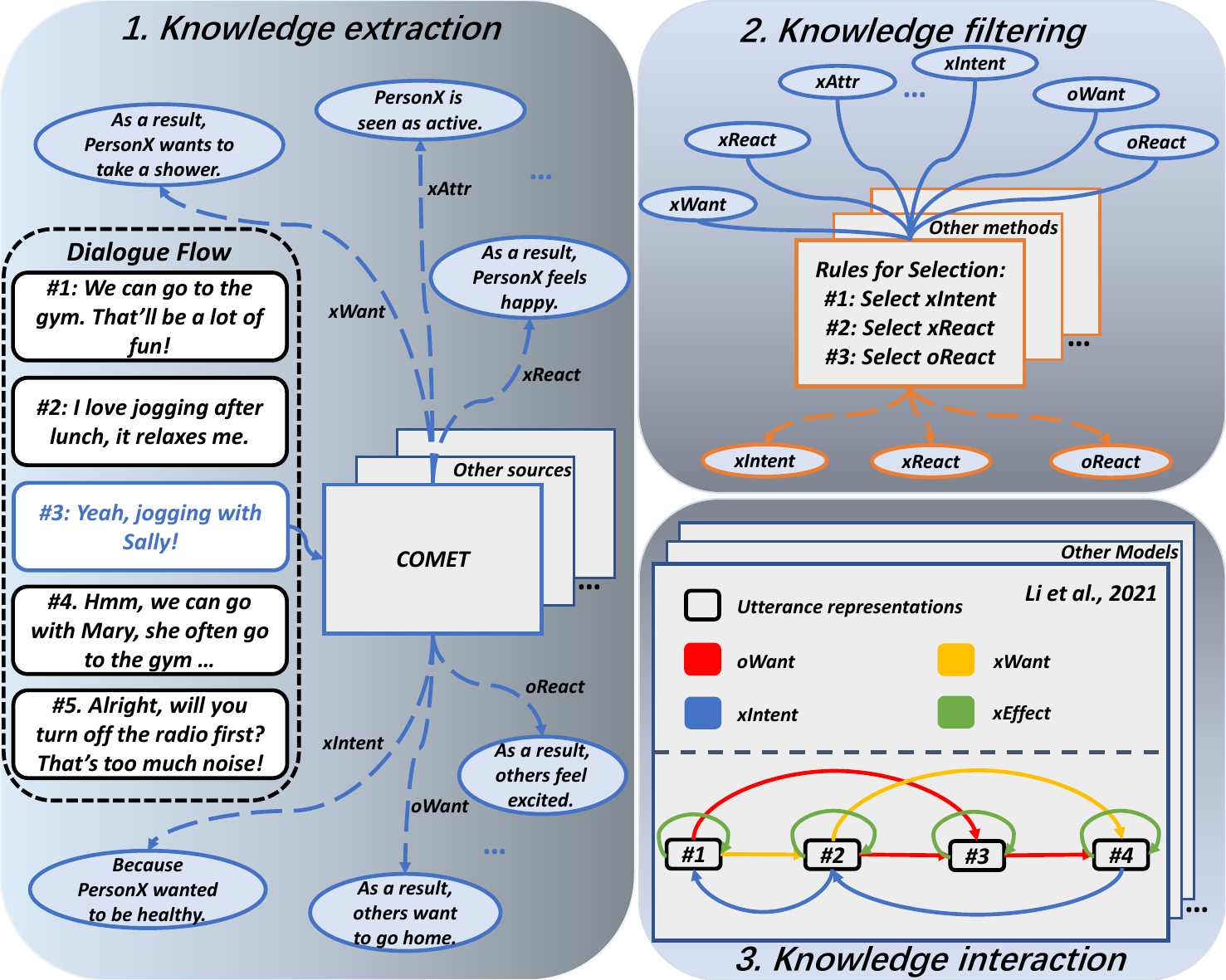}
\caption{Illustration of the three-step paradigm of knowledge infusion. An example based on the knowledge source COMET is provided for each step.}
\label{fig:example}
\end{figure}

There are two key challenges for enhancing conversational emotional reasoning. Firstly, the emotion of the target speaker is influenced by both his own mental state and other participants' behaviors. Current methods mainly build conversation models~\citep{shen2021dialogxl,yang2023cluster,shen-etal-2021-directed} based on Pre-trained Language Models (PLMs)~\citep{liu2019roberta,yang2019xlnet} to tackle these dependencies. Secondly, emotions are often conveyed implicitly with metaphor, sarcasm, and underlying common sense. To mine the related information, a mainstream solution infuses commonsense knowledge to provide emotional clues and help model the inter-utterance relations, which mostly follows a three-step paradigm. We provide an illustration in Figure \ref{fig:example} and summarize this paradigm as follows: Firstly, \textbf{\emph{knowledge extraction}} obtains commonsense knowledge items with the target conversation as queries. This process is closely dependent on the granularity of the queries and the characteristics of the knowledge source. As shown in the example, with the utterance-level query "Yeah, jogging with Sally!" and the generative knowledge source COMET~\citep{bosselut-etal-2019-comet}, we can extract sentence-level social commonsense knowledge such as "PersonX is seen as active" and "As a result, others feel excited". Secondly, \textbf{\emph{knowledge filtering}} selects the most relevant knowledge aspects from the extracted knowledge items based on task-specific priors. The most widely adopted methods include rule-based~\citep{ghosal-etal-2020-cosmic,zhu-etal-2021-topic,zhao2022knowledge} and distantly supervised~\citep{DBLP:conf/ijcai/LiM0L0C0Z22,yang2023evaluations} filtering. Finally, \textbf{\emph{knowledge interaction}} introduces the filtered knowledge to the conversation model via customized architectures to enhance its emotional reasoning ability. For example, a typical architecture~\citep{li-etal-2021-past-present,DBLP:conf/ijcai/LiM0L0C0Z22} assigns knowledge aspects to the edges of the conversation graphs and utilizes the corresponding knowledge features as edge representations to model the inter-utterance dependencies.

In this paper, we dive into the knowledge filtering and knowledge interaction steps and raise the following research questions (RQ):
\begin{itemize}
    \item \textbf{RQ1}: Most knowledge filtering methods select knowledge aspects empirically without further evaluations. What knowledge aspects are most effective in enhancing conversational emotional reasoning?
    \item \textbf{RQ2}: Current knowledge interaction methods are highly coupled with both the knowledge sources and conversation models. Are these customized architectures necessary?
\end{itemize}
In addressing these research questions, we propose a simple yet effective \textbf{B}ipartite \textbf{H}eterogeneous \textbf{G}raph (BHG) method for enhancing emotional reasoning with commonsense knowledge. Firstly, we extract context-aware utterance representations via a PLM-based conversation model and relevant commonsense knowledge from three multi-type and multi-grained knowledge sources. Considering the complementary nature~\citep{DBLP:conf/ijcai/0002WZNG0CG21} between utterance semantics and related common sense, we model the extracted utterance and knowledge representations as heterogeneous nodes. In addition, we introduce a forward and a backward knowledge aggregation node type to perform automatic knowledge filtering and knowledge interaction, because for each target utterance, effective knowledge aspects for modeling inter-utterance relations are usually different in the past context and future context along the dialogue flow, as shown in previous works~\citep{li-etal-2021-past-present,DBLP:conf/ijcai/LiM0L0C0Z22,zhao2022knowledge}. Then a bipartite graph is built on these four heterogeneous node types, where messages from the utterance and knowledge are passed to the knowledge aggregation nodes for semantic-aware knowledge filtering, and the aggregated knowledge messages interact with the utterance nodes to enrich their semantics and model inter-utterance dependencies. The BHG is decoupled from the conversation models and knowledge sources and enables a unified model architecture for multi-type and multi-grained knowledge infusion. Its simple bipartite structure also facilitates the graph reasoning process.

In a BHG, heterogeneous nodes usually possess different feature spaces with unequal dimensions as conversation models/knowledge sources change. During the graph reasoning process, existing heterogeneous graph neural networks~\citep{schlichtkrull2018modeling,wang2019heterogeneous,DBLP:conf/www/HuDWS20} mostly project all types of nodes into a unified feature space to facilitate the interaction between neighbors. However, the projections disrupt the original feature spaces and can lead to unnecessary loss of information for high-dimensional node types. Based on the heterogeneous graph Transformer~\citep{DBLP:conf/www/HuDWS20}, we propose a Multi-dimensional Heterogeneous Graph Transformer (MHGT) for graph reasoning, which utilizes a multi-dimensional edge-dependent matrix to enable direct attention calculation and message passing between heterogeneous nodes with different feature dimensions. MHGT allows all node types to retain the original feature spaces and potentially useful information for the knowledge filtering and knowledge interaction processes during inference.

We evaluate the effectiveness of our proposed BHG and MHGT methods on five datasets across two conversational emotional reasoning tasks: Emotion Recognition in Conversations (ERC) and Casual Emotion Entailment (CEE). The experimental results show that the BHG-based methods outperform previous state-of-the-art knowledge infusion models, and show generalized knowledge infusion ability on multi-type and multi-grained knowledge sources with higher efficiency than previous customized methods. We also analyze the effectiveness of different knowledge aspects, and the results show that previous empirical knowledge filtering methods can introduce less useful knowledge aspects and discard knowledge aspects that benefit the emotional reasoning process.

In summary, this paper makes the following contributions: (1) we propose a bipartite heterogeneous graph-based method to enhance emotional reasoning with commonsense knowledge, which enables a unified framework for multi-type and multi-grained knowledge infusion; (2) we propose a multi-dimensional heterogeneous graph Transformer for graph reasoning, which allows unchanged feature spaces and unequal dimensions for heterogeneous node representations during inference; (3) the BHG-based methods significantly outperform state-of-the-art knowledge infusion methods and show generalized knowledge infusion ability with higher efficiency.

\section{Methodology}
We introduce the utterance and knowledge feature extraction process in Sec. \ref{fe}. Then the BHG construction process and the multi-dimensional HGT methods are introduced in Sec. \ref{BHG} and \ref{MHGT}. Finally, the examined tasks and prediction process are described in Sec. \ref{task}.

\begin{figure*}[htpb]
\centering
\includegraphics[width=15cm,height=6cm]{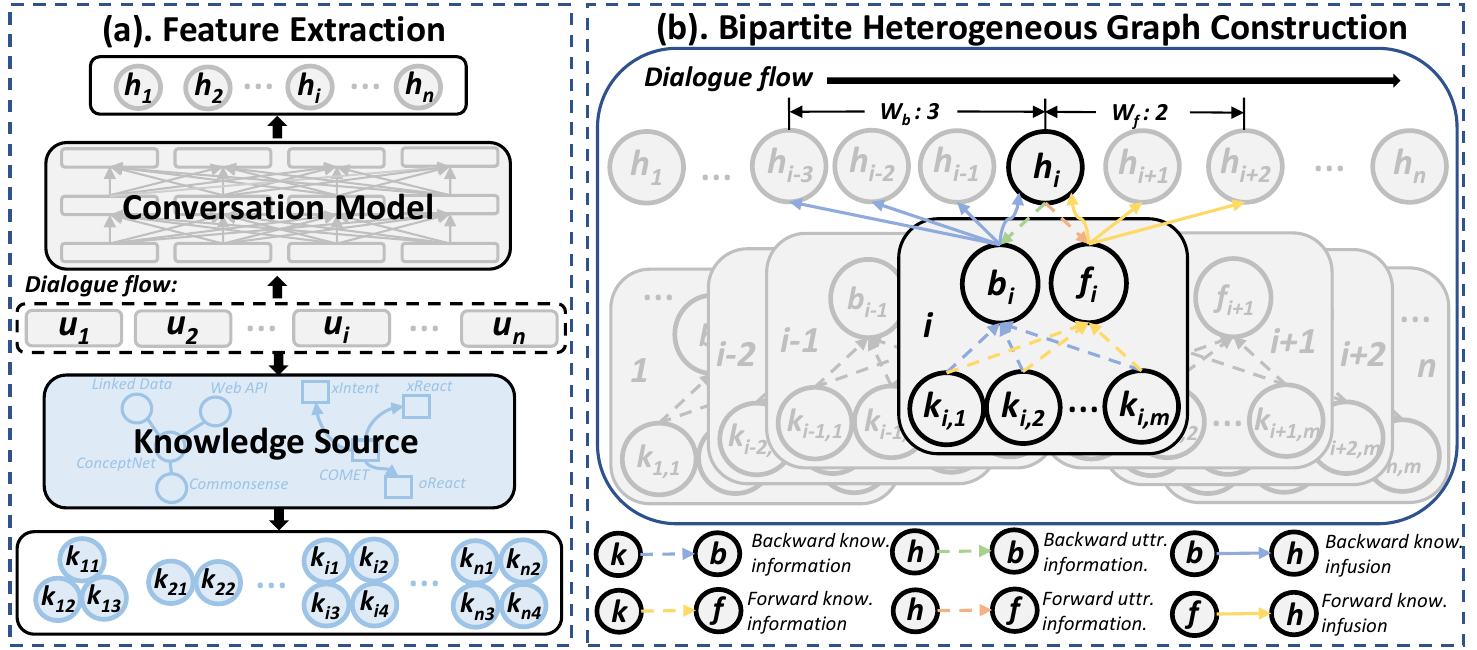}
\caption{An overview of the utterance/knowledge feature extraction and BHG construction processes. In (b), the graph construction process for the $i$-th utterance is presented. $W_f=2$ and $W_b=3$ are examples of the forward and backward knowledge infusion window sizes. \emph{know.} and \emph{uttr.} denote "knowledge" and "utterance".}
\label{fig:model}
\end{figure*}

\subsection{Feature Extraction}\label{fe}
In this section, we extract the utterance features via a PLM-based conversation model. We also extract multi-type (generative and extractive knowledge) and multi-grained (utterance-level and phrase-level) commonsense knowledge to enhance the emotional reasoning process. These processes are illustrated in Fig. \ref{fig:model} (a).

\subsubsection{Conversation Model}
We decouple conversation modeling from knowledge infusion by utilizing existing conversation models to obtain context-aware utterance representations. We utilize a PLM-based conversation model. For a conversation $\mathcal{D}=[$$u_1$, $u_2$, \dots, $u_i$, \dots, $u_N$$]$ with $N$ utterances, $u_i$ is the target utterance pre-pended with its speaker. The model concatenates both past and future contexts as the input $\hat{u}_i=\mathop{||}_{j\in[1,~N]}u_j$, where $\mathop{||}$ is the concatenation operation. Considering situations where only the dialogue history is available, we also test our methods on only past contexts: $\hat{u}_i=\mathop{||}_{j\in[1,~i]}u_j$. We use a RoBERTa-Large~\citep{liu2019roberta} encoder to model the dialogue:
\begin{equation}
    \hat{h}_i = \operatorname{RoBERTa}(\hat{u}_i)
\end{equation}
where $\hat{h}_i\in\mathbb{R}^{N_t\times D_h}$ denotes the token-level representations, $N_t$ is the token number of $\hat{u}_i$, and $D_h$ denotes the dimension of hidden states. We further obtain the utterance-level context-aware representation $h_j$ for each utterance $u_j$ within the dialogue flow via mean-pooling:
\begin{equation}
    h_j = \operatorname{MeanPooling}(\hat{h}_i[l_j:r_j])
\end{equation}
where $l_j$ and $r_j$ denote the start and end positions of $u_j$ in $\hat{u}_i$.

\subsubsection{Knowledge Extraction}
In previous works, two commonsense knowledge graphs have been proven most effective in enhancing the emotional reasoning process: the social commonsense knowledge graph ATOMIC~\citep{sap2019atomic} and the taxonomic/lexical knowledge graph ConceptNet~\citep{speer2017conceptnet}. Therefore, we separately obtain relevant knowledge from three knowledge sources expanded from ATOMIC and ConceptNet.

With the development of automatic knowledge graph construction, generative knowledge sources receive increasing interest due to their flexibility and convenience in knowledge extraction. We extract the ATOMIC knowledge via two COMET models: the first one is COMET$_{2019}$~\citep{bosselut-etal-2019-comet}, which pre-trains a GPT~\citep{radford2018improving} model on ATOMIC. During knowledge extraction, each utterance $u_i$ is constructed into the query: $(u_i\mathop{||}a_j\mathop{||}[GEN])$. $a_j$ denotes a knowledge aspects term representing an if-then relation type for the speaker's actions/mental states. COMET$_{2019}$ provides nine knowledge aspects from ATOMIC, where their interpretations are listed in Table \ref{tab:aspect}. The constructed query is then input to the COMET model, and the final-layer hidden representation of the decoder: $k_{ij}\in\mathbb{R}^{D_k^{2019}}$ is used as the utterance-level knowledge representation corresponding to $a_j$, where $D_k^{2019}$ denotes the dimension of $k_{ij}$. 

\begin{table}[h]
\caption{Interpretations of the selected knowledge aspects/relations. "X" denotes the target utterance speaker.}
\label{tab:aspect}
\begin{center}
\setlength{\tabcolsep}{0.5mm}{
\begin{tabular}{lc}
\toprule \multicolumn{2}{c}{\textbf{COMET}}\\ \midrule
\textbf{Aspect} &  \textbf{Interpretation}\\ \midrule
\emph{xIntent} & Why does X cause the event?\\
\emph{xAttr} & How would X be described?\\
\emph{xNeed} & What does X need to do before the event?\\
\emph{xWant} & What would X likely want to do after the event?\\
\emph{xEffect} & What effects does the event have on X?\\
\emph{xReact} & How does X feel after the event?\\
\emph{oWant} & What would others likely want to do after the event?\\
\emph{oEffect} & What effects does the event have on others?\\
\emph{oReact} & How do others feel after the event?\\ \midrule
\multicolumn{2}{c}{\textbf{ConceptNet}}\\ \midrule
\textbf{Relation} &  \textbf{Interpretation}\\ \midrule
\emph{IsA} & is a\\
\emph{HasProperty} & has the property of\\
\emph{DefinedAs} & is defined as\\
\emph{AtLocation} & is located at\\
\emph{RelatedTo} & is related to\\
\bottomrule
\end{tabular}}
\end{center}
\end{table}

The second knowledge source COMET$_{2020}$\citep{DBLP:conf/aaai/HwangBBDSBC21} further extends the generative pre-training to an expanded ATOMIC knowledge graph and part of ConceptNet based on a larger BART~\citep{lewis-etal-2020-bart} model. COMET$_{2020}$ follows a similar knowledge extraction process as COMET$_{2019}$ and each knowledge representation has dimension $D_k^{2020}$. 

On the other hand, we obtain phrase-level ConceptNet knowledge in an extractive manner. Specifically, we tokenize each utterance $u_i$ and concatenate the tokens into n-gram phrases. For each phrase, we extract all its immediate neighbors in the English sub-graph of ConceptNet. Each neighbor assertion is a $\langle$\emph{source}, \emph{relation}, \emph{target}, \emph{weight}$\rangle$ quadruple where \emph{source} and \emph{target} denote the query phrase and the neighbor concept, \emph{relation} denotes the corresponding relation type, and \emph{weight} denotes a confidence score assigned to the assertion. For example, we can extract the assertion: $\langle$\emph{think}, \emph{HasPrerequisite}, \emph{use brains}, 2.375$\rangle$ for the input phrase \emph{think}. There are 28 different relation types in the extracted quadruples. We further remove all assertions with confidence scores less than 2.0 for denoising. To facilitate the extraction of the knowledge representations, we manually design an interpretation for each relation type and convert all assertions into natural language by concatenating the concepts to the relation interpretation. For example, $\langle$\emph{think}, \emph{HasPrerequisite}, \emph{use brains}$\rangle$ is converted to "think has the prerequisite of use brains". We present the interpretations of the five most common relations in Table \ref{tab:aspect}. As most of these knowledge items have simple semantics, we use a RoBERTa-Base encoder to extract the features for the converted natural language knowledge and each knowledge representation has dimension $D_k^{Concept}$.

Overall, for each utterance $u_i$ and a knowledge source, we obtain a set of knowledge representations: $\{k_{i1},k_{i2},...,k_{im_i}\}$, where $k_{ij}\in\mathbb{R}^{D_k}$, the representation dimension $D_k\in\{D_k^{2019}, D_k^{2020}, D_k^{Concept}\}$ depends on the knowledge source, and $m_i$ denote the number of extracted knowledge items for $u_i$.

\subsection{Bipartite Heterogeneous Graph}\label{BHG}
We propose a directed bipartite heterogeneous graph (BHG) to complement utterance representations with extracted commonsense knowledge representations. Formally, the BHG is denoted as $\mathcal{G}=(\mathcal{V}, \mathcal{E}, \mathcal{N}, \mathcal{R})$, where each node $v\in\mathcal{V}$ and each edge $e\in\mathcal{E}$. $\mathcal{N}, \mathcal{R}$ denote the set of node and relation types, and each node or edge is projected to its type via a mapping function: $\gamma(v):\mathcal{V}\rightarrow\mathcal{N}$, $\tau(e):\mathcal{E}\rightarrow\mathcal{R}$, respectively. Specifically, the node and edge types are defined as follows:

\subsubsection{Node Types}
Firstly, each utterance within a conversation is modeled as a node in the BHG. Its representation obtained from the conversation model is used as the node feature. The utterance node type is denoted as $h$. The extracted knowledge representations are modeled as another type of node: $k$. In addition, we introduce extra knowledge aggregation node types to perform automatic knowledge filtering and knowledge interaction. As previous works have shown that effective knowledge aspects can be different for performing knowledge interactions with the past context and future context~\citep{li-etal-2021-past-present,DBLP:conf/ijcai/LiM0L0C0Z22,zhao2022knowledge}, we introduce a forward and a backward aggregation node type, which is separately denoted as $f\in\mathbb{R}^{D_f}$ and $b\in\mathbb{R}^{D_b}$, where $D_f$ and $D_b$ are their representation dimensions. Overall, we define a set with four types of nodes: $\mathcal{N}=\{h, k, f, b\}$.

\subsubsection{Relation Types}
For each edge $e=\langle v_1, v_2\rangle$ from the source node $v_1$ to the target node $v_2$, its relation type is defined as $\tau(e): \langle\gamma(v_1), \gamma(v_2)\rangle$. Specifically, we introduce six types of relations to perform knowledge filtering and knowledge interaction. Firstly, the forward/backward knowledge information relations: $\tau_{kf}:\langle k, f\rangle$ and $\tau_{kb}:\langle k, b\rangle$ are proposed to separately provide the extracted knowledge information to both forward and backward knowledge aggregation nodes. Secondly, the knowledge filtering process naturally requires considering the corresponding utterance information. Therefore, we utilize the utterance information relations: $\tau_{uf}:\langle h, f\rangle$ and $\tau_{ub}:\langle h, b\rangle$ to introduce utterance semantics to the knowledge aggregation nodes. In addition, commonsense knowledge has been proven not only useful for enriching the semantics of each utterance but in modeling inter-utterance dependencies~\citep{xie-etal-2021-knowledge-interactive,zhao2022knowledge,DBLP:conf/ijcai/ZhaoZL22}. Therefore, we further incorporate the filtered commonsense knowledge to the utterance nodes by designing a forward infusion relation: $\tau_{f}:\langle f, h\rangle$ and a backward infusion relation: $\tau_{b}:\langle b,h\rangle$. Overall, the relation set contains six types of relations: $\mathcal{R}=\{\tau_{kf},\tau_{kb},\tau_{uf},\tau_{ub},\tau_{f},\tau_{b}\}$.

\subsubsection{BHG Construction}
For each utterance node $h_{i}$, we create a forward aggregation node $f_i$ and a backward aggregation node $b_i$. Then we build a BHG for each conversation by considering the following criteria for each relation type:
\begin{itemize}
    \item $\forall i\neq j$ and $m,\langle k_{im},f_{j}\rangle\notin\mathcal{E}$ and $\langle k_{im},b_{j}\rangle\notin\mathcal{E}$. Each knowledge aggregation node only receives and filters the extracted knowledge from its corresponding utterance.
    \item $\forall i\neq j,\langle h_i,f_j\rangle\notin\mathcal{E}$ and $\langle h_i,b_j\rangle\notin\mathcal{E}$. Each knowledge aggregation node only considers the information from its corresponding utterance when performing knowledge filtering.
    \item $\forall i\leq j\leq i+W_f,\langle f_i,h_j\rangle\in \mathcal{E}$, where $W_f$ is the pre-defined forward knowledge infusion window size. Each forward aggregated knowledge is only used to enrich the semantics of its own utterance and model the inter-utterance relations with future utterances.
    \item $\forall i-W_b\leq j \leq i,\langle b_i,h_j\rangle\in \mathcal{E}$, where $W_b$ is the pre-defined backward knowledge infusion window size. Each backward aggregated knowledge is only used to enrich the semantics of its own utterance and model the inter-utterance relations with previous utterances.
\end{itemize}
Based on these criteria, the BHG building process is described in Algorithm \ref{algorithm-bhg}, and an example is illustrated in Figure \ref{fig:model}(b).

\begin{algorithm} 
	\caption{Construction of the BHG} 
	\label{algorithm-bhg} 
	\begin{algorithmic}[1]
		\Require dialogue representations: $\mathcal{H}$ = [$h_1$, $h_2$,..., $h_n$], knowledge representations: $\mathcal{K}$ = [$k_{11}$,..., $k_{1m_1}$,..., $k_{n1}$,..., $k_{nm_n}$], forward and backward knowledge infusion window sizes: $W_f$, $W_b$
		\Ensure the BHG $\mathcal{G}=(\mathcal{V}, \mathcal{E}, \mathcal{N}, \mathcal{R})$
            \State $\mathcal{V} \gets \mathcal{H}\cup\mathcal{K}$, $\mathcal{E} \gets \varnothing$ \Comment{Initialize the BHG.}
            \State $\mathcal{N} \gets \{h, k, f, b\}$ 
            \State $\mathcal{R} \gets \{\tau_{kf},\tau_{kb},\tau_{uf},\tau_{ub},\tau_{f},\tau_{b}\}$
		\For{$i \in \{1,...,n\}$}
            \State $\mathcal{V}\gets\mathcal{V}\cup\{f_i,b_i\}$ \Comment{Add the aggregation nodes.}
            \For{$j \in \{1,...,m_i\}$}
            \State $\mathcal{E}\gets\mathcal{E}\cup\{\langle k_{ij},f_i\rangle\}$\Comment{Add knowledge info. relations.}
            \State $\mathcal{E}\gets\mathcal{E}\cup\{\langle k_{ij},b_i\rangle\}$
            \EndFor
            \State $\mathcal{E}\gets\mathcal{E}\cup\{\langle h_i,f_i\rangle\}$\Comment{Add utterance info. relations.}
            \State $\mathcal{E}\gets\mathcal{E}\cup\{\langle h_i,b_i\rangle\}$
            \State $c_f=0$
            \While{$c_f\leq W_f$ and $i+c_f\leq n$}
            \State $\mathcal{E}\gets\mathcal{E}\cup\{\langle f_i,h_{i+c_f}\rangle\}$\Comment{Forward knowledge infusion.}
            \State $c_f\gets c_f + 1$
            \EndWhile
            \State $c_b=0$
            \While{$c_b\leq W_b$ and $i-c_b\geq0$}
            \State $\mathcal{E}\gets\mathcal{E}\cup\{\langle b_i,h_{i-c_b}\rangle\}$\Comment{Backward knowledge infusion.}
            \State $c_b\gets c_b + 1$
            \EndWhile
            \EndFor 
	\end{algorithmic} 
\end{algorithm}

\subsection{Multi-dimensional HGT}\label{MHGT}
In heterogeneous graphs, different types of nodes/relations do not share feature spaces, which requires node and relation-dependent architectures for GNN-based graph modeling. The Heterogeneous Graph Transformer (HGT)~\citep{DBLP:conf/www/HuDWS20} separately initializes and optimizes a set of Transformer~\citep{vaswani2017attention} parameters for each homogeneous sub-graph. In addition, HGT unifies the dimensions of all node types before aggregation via linear transformation due to the complicated structures of most large-scale heterogeneous graphs. In our case, the utterance and knowledge representations often have unequal dimensions: $D_h\neq D_k$ with the substitution of conversation models and knowledge sources. Considering the simple structure of BHG, we believe unifying dimensions can lead to unnecessary loss of information for high-dimensional node types. Therefore, we propose a Multi-dimensional HGT (MHGT) to model the BHG, which allows the dimensions of heterogeneous nodes to remain unchanged during the attention calculation and message-passing processes. 

Based on the structure of vanilla HGT, MHGT improves the heterogeneous dot-product attention process to enable interactions between nodes with different dimensions. For a target node representation $v_t^l\in\mathbb{R}^{D_{\gamma(v_t)}}$ at the $l$-th MHGT layer, we calculate the multi-head attention weights for its neighbor $v_n^l\in\mathbb{R}^{D_{\gamma(v_n)}}$ as follows:
\begin{small}
\begin{eqnarray}
Att\_head^i(v_n^l,v_t^l) = \frac{K^i_{\gamma(v_n)}(v_n^l)W_{\tau(\langle v_n,v_t\rangle)}^{Att}Q^i_{\gamma(v_t)}(v_t^l)}{\sqrt{D_{\gamma(v_t)}}} \\
Att(v_n^l, v_t^l) = \mathop{Softmax}\limits_{\langle v_n,v_t\rangle\in\mathcal{E}}\left(\mathop{||}\limits_{i\in[1,h]}Att\_head^i(v_n^l,v_t^l)\right)
\end{eqnarray}
\end{small}
where $K^i_{\gamma(v_n)}(v_n)\in\mathbb{R}^{\frac{D_{\gamma(v_n)}}{h}}$ and $Q^i_{\gamma(v_t)}(v_t)\in\mathbb{R}^{\frac{D_{\gamma(v_t)}}{h}}$ denotes the $i$-th head of a equal-dimensional linear projection from $v_n^l$ as the key and $v_t^l$ as the query, $||$ denotes concatenation, and $Softmax$ denotes the softmax operation. Different from the vanilla HGT, MHGT proposes a multi-dimensional edge-dependent matrix $W_{\tau(\langle v_n,v_t\rangle)}^{Att}\in\mathbb{R}^{\frac{D_{\gamma(v_n)}}{h}\times\frac{D_{\gamma(v_t)}}{h}}$ to project nodes of $\gamma(v_n)$ to the representation space of $\gamma(v_t)$, which enables attention calculations between heterogeneous node features and allows situations where $D_{\gamma(v_n)}\neq D_{\gamma(v_t)}$. We perform the message-passing process for $v_n^l$ with the target node $v_t^l$ as follows:
\begin{eqnarray}
Mes\_head^i(v_n^l,v_t^l) = V^i_{\gamma(v_n)}(v_n^l)W_{\tau(\langle v_n,v_t\rangle)}^{Mes} \\
Mes(v_n^l, v_t^l) = \mathop{||}\limits_{i\in[1,h]}Mes\_head^i(v_n^l,v_t^l)
\end{eqnarray}
where $V_{\gamma(v_n)}(v_n)$ is another equal-dimensional linear projection from $v_n$ as the value. Similarly, MHGT uses a multi-dimensional edge-dependent matrix $W_{\tau(\langle v_n,v_t\rangle)}^{Mes}\in\mathbb{R}^{\frac{D_{\gamma(v_n)}}{h}\times\frac{D_{\gamma(v_t)}}{h}}$ to align the neighbor node $v_n$ to the feature space of the target node $v_t$. Same as in vanilla dot-product attention, the message is aggregated using the calculated attention weights:
\begin{small}
    \begin{equation}
    \hat{v}_t^{l+1} = \mathop{\oplus}\limits_{\langle v_n,v_t\rangle\in\mathcal{E}}\left(Att(v_n^l, v_t^l)\cdot Mes(v_n^l, v_t^l)\right)
    \end{equation}
\end{small}
where $\oplus$ denotes the element-wise sum operation. Finally, $\hat{v}_t^{l+1}$ is used to update $v_t^l$ in the following manner:
\begin{equation}
    v_t^{l+1} = T_{\gamma(v_t)}(\sigma(\hat{v}_t^{l+1})) + v_t^{l}
\end{equation}
where $\sigma$ denotes the $Gelu$ activation function and $T_{\gamma(v_t)}$ denotes an equal-dimensional linear projection from $\gamma(v_t)$. We stack $L$ MHGT layers to allow interactions between non-adjacent nodes. For utterance $u_i$, we obtain the $L$-th layer output from MHGT: $h_i^L$ as its final knowledge-enhanced representation.

\subsection{Prediction and Training}\label{task}
We examine the BHG-based knowledge infusion method on two complex emotional reasoning tasks: Emotion Recognition in Conversations (ERC) and Casual Emotion Entailment (CEE). 

\subsubsection{Emotion Recognition in Conversations}
ERC aims to identify each utterance $u_i$'s emotion within a dialogue $\mathcal{D}$ from a pre-defined emotion category set $E$, which is modeled as a text classification task on each utterance~\citep{DBLP:journals/access/PoriaMMH19}. Specifically, we utilize a feed-forward neural network to project the knowledge-enhanced utterance representations to the classification space:
\begin{equation}\label{erc_predict}
    \hat{y}_i = Softmax(h_i^LW_{ERC}+b_{ERC})
\end{equation}
where $W_{ERC}\in\mathbb{R}^{D_h\times|E|}$ and $b_{ERC}\in\mathbb{R}^{|E|}$ are learnable parameters. We optimize the standard cross-entropy loss to train the ERC model:
\begin{equation}\label{erc_loss}
    \mathcal{L}_{ERC} = -\frac{1}{N}\sum_{i=1}^N\sum_{j=1}^{|E|}y_{ij}log(\hat{y}_{ij})
\end{equation}
where $\hat{y}_{ij}$ and $y_{ij}$ denote the $j$-th element of $\hat{y}_i$ and the one-hot emotion label $y_i$ of utterance $u_i$, and $N$ denotes the batch size.

\subsubsection{Casual Emotion Entailment}
CEE aims to identify the causes behind the non-neutral emotion of target utterances and locate their positions from the conversational history. Given a dialogue $\mathcal{D}$ and the emotion label $y$ of each utterance, CEE is modeled as a binary classification task to predict whether the candidate utterance $u_j$ contains the emotion cause for target utterance $u_i$, where $1\leq j\leq i$. During inference, we concatenate the knowledge-enhanced utterance representations of $u_j$ and $u_i$ to calculate the logits:
\begin{equation}
    \hat{z}_{ji} = \frac{1}{1+e^{-([h_j^L;h_i^L]W_{CEE}+b_{CEE})}}
\end{equation}
where $W_{CEE}\in\mathbb{R}^{2D_h\times1}$ and $b_{CEE}\in\mathbb{R}^1$ are learnable parameters, respectively. A BCE loss is used to incorporate the supervision signals:
\begin{equation}
    \mathcal{L}_{BCE} = -\frac{1}{N}\sum_{i=1}^N\sum_{j=1}^{i}z_{ji}log(\hat{z}_{ji})
\end{equation}
where $z_{ji}\in\{0,1\}$ is the binary label. We further introduce the emotion labels to enhance the learning of the CEE model using the same prediction and training paradigm as in Eqn. \ref{erc_predict} and Eqn. \ref{erc_loss}, and jointly optimize the two tasks in a multi-task learning manner:
\begin{equation}
    \mathcal{L}_{CEE} = \mathcal{L}_{BCE}+\alpha\mathcal{L}_{ERC}
\end{equation}
where $\alpha$ denotes a hyper-parameter controlling the weight of the ERC loss.


\section{Experiments}
\subsection{Datasets}
We test our method on four Emotion Recognition in Conversations (ERC) datasets and one Causal Emotion Entailment (CEE) dataset. For all datasets, we only utilize the text modality in our experiments. 

\textbf{IEMOCAP}~\citep{DBLP:journals/lre/BussoBLKMKCLN08}: A two-party multi-modal ERC
dataset derived from the scenarios in the scripts of the two actors. The pre-defined emotion category set $E$ consists of: \emph{neutral, sad, anger, happy, frustrated, excited}.

\textbf{MELD}~\citep{poria-etal-2019-meld}: A multi-party multi-modal ERC dataset collected from the scripts of American TV show \emph{Friends}. The pre-defined emotions are \emph{neutral, sad, anger, disgust, fear, happy, surprise}.

\textbf{DailyDialog}~\citep{li-etal-2017-dailydialog}: A ERC dataset compiled from human-written daily conversations with only two parties involved and no speaker information. The pre-defined emotion labels are \emph{neutral, happy, surprise, sad, anger, disgust, fear}.

\textbf{EmoryNLP}~\cite{zahiri2017emotion}: Another ERC dataset collected from TV show \emph{Friends}. It is annotated with the following emotion categories: \emph{neutral, sad, mad, scared, powerful, peaceful, joyful}.

\textbf{RECCON}~\citep{DBLP:journals/cogcom/PoriaMHGBJHGRCG21}: A CEE dataset collected from the scripts of DailyDialog with both utterance-level emotion labels and binary emotion cause labels, with the same emotion category set as DailyDialog.

\subsection{Baseline Models}
We compare our method with strong baselines models and categorize them into four groups according to their characteristics:

\paragraph{PLM-based methods} 
For the ERC task, we select the following two methods: \textbf{RoBERTa-Large}~\citep{liu2019roberta} used the PLM RoBERTa-Large to directly model the conversation. The utterance representations are used to fine-tune the weights. \textbf{DialogXL}~\citep{shen2021dialogxl} improved the XLNet~\citep{yang2019xlnet} with the enhanced memory and dialog-aware self-attention mechanism to capture long historical context and dependencies between multiple parties. For the CEE task, we select: \textbf{RoBERTa-Base/Large}~\citep{DBLP:journals/cogcom/PoriaMHGBJHGRCG21} concatenated the conversation as input to the PLM RoBERTa. Then CEE was modeled as a binary classification problem for each utterance pair.

\paragraph{Graph-based Methods}
For ERC, \textbf{RGAT}~\cite{ishiwatari-etal-2020-relation} improved the relation modeling of conversations and added relational position encodings as sequential information. \textbf{DAG-ERC}~\citep{shen-etal-2021-directed} built a directed acyclic graph on the dialogue and used a graph neural network to aggregate the information. For CEE, we select two methods: \textbf{ECPE-2D}~\citep{ding-etal-2020-ecpe} represented the emotion-cause pairs as 2D representations and utilized the Transformer to model them. \textbf{RankCP}~\citep{wei-etal-2020-effective} ranked the clause pairs and performed end-to-end extraction with inter-clause modeling.

\paragraph{Knowledge-based Methods}
For ERC, five methods are selected: \textbf{KI-Net}~\citep{xie-etal-2021-knowledge-interactive} infused both commonsense and sentiment lexicon knowledge and proposed a self-matching module to enhance the knowledge interaction. \textbf{COSMIC}~\citep{ghosal-etal-2020-cosmic} used the RNN to model the dialogue history and extracted utterance-level commonsense knowledge to model the speakers’ mental states. \textbf{TODKAT}~\citep{zhu-etal-2021-topic} modeled topic information via PLMs and explicitly infused event-centered knowledge. \textbf{SKAIG}~\citep{li-etal-2021-past-present} extracted psychological commonsense knowledge and infused the knowledge to enhance the edge representations of the knowledge graph. \textbf{CauAIN}~\citep{DBLP:conf/ijcai/ZhaoZL22} used the emotion cause knowledge to guide the traceback process of context modeling. For CEE, \textbf{KAG}~\citep{yan-etal-2021-position} utilized the entity-related commonsense knowledge to model the semantic dependencies between the candidates and emotions. \textbf{AKM}~\citep{turcan-etal-2021-multi} combined knowledge via an adapted knowledge model in a multi-task learning manner. \textbf{KEC}~\citep{DBLP:conf/ijcai/LiM0L0C0Z22} utilized the directed acyclic graph incorporating social commonsense knowledge to improve the causal reasoning ability. \textbf{KBCIN}~\citep{zhao2022knowledge} proposed the knowledge-bridged causal interaction network to capture context dependencies of conversations and make emotional cause reasoning.

\paragraph{Zero-shot Method with ChatGPT}. We also include the zero-shot evaluation results of the latest large language model ChatGPT~\footnote{\url{https://openai.com/blog/chatgpt}} on all datasets, provided by \citet{yang2023evaluations}.

\begin{table*}[h]
\caption{Test results of our BHG-based methods and baseline models on the five datasets. "Know. Source" lists the commonsense knowledge sources used by each knowledge-based method. In "w/o future context", only the dialogue history is introduced as context for each target utterance. The "w/o emotions" setting removes the utterance emotion information for CEE. Best values are highlighted in bold.}\label{overall_results}
\label{tab:results}
\begin{center}
\resizebox{.9\textwidth}{!}{
\begin{tabular}{lcccccc|lcccc}
\toprule \multicolumn{2}{c}{\textbf{ERC}} & \textbf{IEMOCAP} & \textbf{MELD} & \multicolumn{2}{c}{\textbf{DailyDialog}} & \textbf{EmoryNLP} & \multicolumn{2}{c}{\textbf{CEE}} & \multicolumn{3}{c}{\textbf{RECCON}}\\ \midrule
Model& Know. Source & Weighted F1 & Weighted F1 & Micro F1 & Macro F1 & Weighted F1 & Model & Know. Source & Neg. F1 & Pos. F1 & Macro F1\\ \midrule
\multicolumn{12}{c}{\textbf{Zero-shot Methods}}\\
ChatGPT$_{ZS}$~\citep{yang2023evaluations} & -- & 53.35 & 61.18 & 43.27 & 38.19 & 32.64 & ChatGPT$_{ZS}$~\citep{yang2023evaluations} & -- & 67.18 & 51.35 & 59.26\\\midrule
\multicolumn{12}{c}{\textbf{PLM-based Methods}}\\
RoBERTa-Large~\citep{liu2019roberta} & \multirow{2}{*}{--} & 55.67 & 62.75 & 55.16 & 48.2 & 37.0 & RoBERTa-Base~\citep{DBLP:journals/cogcom/PoriaMHGBJHGRCG21} & \multirow{2}{*}{--} & 88.74 & 64.28 & 76.51\\
DialogXL~\citep{shen2021dialogxl} &  & 65.94 & 62.41 & 54.93 & -- & 34.73 & RoBERTa-Large~\citep{DBLP:journals/cogcom/PoriaMHGBJHGRCG21} &  & 87.89 & 66.23 & 77.06\\ \midrule
\multicolumn{12}{c}{\textbf{Graph-based Methods}}\\
RGAT~\cite{ishiwatari-etal-2020-relation} & \multirow{2}{*}{--} & 66.36 & 62.8 & 59.02 & -- & 34.42 & ECPE-2D~\citep{ding-etal-2020-ecpe} & \multirow{2}{*}{--} & 94.96 & 55.50 & 75.23\\
DAG-ERC~\citep{shen-etal-2021-directed} & & 68.03 & 63.65 & 59.33 & -- & 39.02 & RankCP~\citep{wei-etal-2020-effective} &  & \textbf{97.30} & 33.00 & 65.15\\ \midrule
\multicolumn{12}{c}{\textbf{Knowledge-based Methods}}\\
CauAIN~\citep{DBLP:conf/ijcai/ZhaoZL22} & $COMET_{2019}$ & 67.61 & 65.46 & 58.21 & 53.85 & -- & KAG~\citep{yan-etal-2021-position} & $ConceptNet$ & 86.35 & 58.18 & 72.26\\
KI-Net~\citep{xie-etal-2021-knowledge-interactive} & $ConceptNet$ & 66.98 & 63.24 & 57.30 & 50.8 & -- & AKM~\citep{turcan-etal-2021-multi} & $COMET_{2019}$ & 88.18 & 64.53 & 76.36\\
COSMIC~\citep{ghosal-etal-2020-cosmic} & $COMET_{2019}$ & 65.28 & 65.21 & 58.48 & 51.05 & 38.11 & KEC~\citep{DBLP:conf/ijcai/LiM0L0C0Z22} & $COMET_{2020}$ & 88.85 & 66.55 & 77.70\\
TODKAT~\citep{zhu-etal-2021-topic} & $COMET_{2019}$ & 61.33 & 65.47 & 58.47 & 52.56 & 38.69 & KBCIN~\citep{zhao2022knowledge} & \multirow{2}{*}{$COMET_{2020}$} & -- & 67.51 & 78.43\\
SKAIG~\citep{li-etal-2021-past-present} & $COMET_{2019}$ & 66.96 & 65.18 & 59.75 & 51.95 &38.88 & \ w/o emotions & & -- & 64.05 & 76.73\\\midrule
\multicolumn{12}{c}{\textbf{BHG-based Methods}}\\
MHGT & \multirow{2}{*}{$COMET_{2020}$} & \textbf{71.20} & 65.54 & \textbf{62.37} & \textbf{54.11} & \textbf{39.06}  & MHGT & \multirow{2}{*}{$COMET_{2020}$} & 90.34 & \textbf{69.13} & \textbf{79.73}\\
\ w/o future context &  & 70.93 & \textbf{66.3} & 60.41 & 53.29 & 38.85  & \ w/o emotions &  & 89.89 & 68.58 & 79.23\\ \hline
MHGT & \multirow{3}{*}{$COMET_{2019}$} & 71.07 & 65.57 & 61.5 & 53.0 & 38.62  & MHGT & \multirow{3}{*}{$COMET_{2019}$} & 90.26 & 68.85 & 79.55\\
\ w/o future context &  & 70.64 & 65.77 & 60.68 & 52.51 & 38.33  & \ w/o emotions &  &89.95 & 68.49 & 79.22\\
HGT &  & 67.2 & 62.86 & 60.0 & 51.27 & 36.5  & HGT &  & 89.79 & 62.94 & 76.37\\ \hline
MHGT & \multirow{3}{*}{$ConceptNet$} & 70.55 & 65.99 & 61.41 & 52.7 & 38.07  & MHGT & \multirow{3}{*}{$ConceptNet$} & 89.53 & 68.6 & 79.07\\
\ w/o future context &  & 70.68 & 65.59 & 60.2 & 51.74 & 37.38  & \ w/o emotions &  &90.03 & 67.72 & 78.88\\
HGT &  & 67.1 & 64.29 & 59.28&51.09& 36.06  & HGT &  &90.12 & 62.47 & 76.29\\
\bottomrule
\end{tabular}}
\end{center}
\end{table*}

\subsection{Implementation and Evaluation Settings}
We conduct all experiments using a single Nvidia Tesla A100 GPU with 80GB of memory. We initialize the pre-trained weights of RoBERTa and use the tokenization tools provided by Huggingface~\citep{wolf2019huggingface}. We leverage AdamW optimizer~\citep{DBLP:conf/iclr/LoshchilovH19} to train the model. The batch size of experiments on all datasets is set to 16 except in DailyDialog, which is 24. We use a linear warm-up learning rate scheduling of warm-up ratio 20$\%$ and a peak learning rate 1e-5. We set a dropout rate of 0.1 and an L2-regularisation rate of 0.01 to avoid over-fitting. For hyper-parameters, $D_h=D_k^{2020}=1024$, $D_k^{2019}=D_k^{Concept}=768$, $L=3$, and $\alpha$ is set to 0.8. For each knowledge source with dimension $D_k$, we set $D_f=D_b=D_k$ and randomly initialize the forward/backward knowledge aggregation node representations. During knowledge infusion, we set $W_f=W_b=5$ when both past and future contexts are provided, and $W_f=0$, $W_b=10$ when only the past context is available.

For ERC, we select the Weighted F1 score as the evaluation metric for IEMOCAP, MELD, and EmoryNLP. Since “neutral” occupies most of DailyDialog, we utilize the Micro F1 score excluding the "neutral" utterances to reflect the performances in non-neutral emotions, as in previous works~\citep{shen-etal-2021-directed,zhu-etal-2021-topic,li-etal-2021-past-present}. We also calculate Macro F1 scores on all classes for DailyDialog to evaluate the overall performances. For CEE, we report the F1 scores on positive and negative utterances and Macro F1 scores as the overall evaluation. All reported results are averages of five random runs.

\section{Results and Analysis}
\subsection{Main Results}
\subsubsection{Overall Performance}
The performance of our BHG-based methods and all baseline models on the five datasets are presented in Table \ref{overall_results}. Firstly, the zero-shot prompting results show that ChatGPT still bears a huge gap with advanced conversation models and knowledge-based methods in performing emotional reasoning, possibly because these tasks are very subjective even to humans, showing the necessity of exploring few-shot prompting and knowledge
infusion to further calibrate ChatGPT's understanding of these subjective emotion concepts~\citep{yang2023evaluations}. These results also motivate continual research on supervised task-specific methods. Secondly, the knowledge-based methods significantly improve model performance on most datasets compared to the PLM-based and graph-based conversation models. These results empirically prove the effectiveness of commonsense knowledge infusion to emotional reasoning tasks. Thirdly, the BHG-based methods outperform all baseline models
on all ERC and CEE datasets, including the new state-of-the-art performance of 71.2\% on IEMOCAP, 62.37\% (Micro-F1) on DailyDialog, and 79.73\% (Macro-F1) on RECCON. These outstanding performances quantify the advantages of BHG-based knowledge infusion over previous methods.

\subsubsection{Knowledge-based Methods Comparisons}
In the comparison of knowledge-based methods, the unified BHG architecture shows generalized knowledge infusion ability by outperforming all previous highly customized knowledge infusion methods on all three tested knowledge sources. For the phrase-level extractive knowledge source ConceptNet, BHG outperforms KI-Net by at least 2\% on IEMOCAP, MELD, and DailyDialog. It also has an impressive 6.81\% improvement on RECCON compared to KAG. These results show that the BHG structure can effectively interact with multi-grained heterogeneous nodes to infuse knowledge. For the utterance-level generative knowledge source $COMET_{2019}$, BHG possesses a similar advantage over other knowledge-based methods on most datasets, especially on IEMOCAP, DailyDialog, and RECCON. A possible reason is that some useful knowledge is mistakenly discarded by these previous methods as they all perform empirical knowledge filtering. On the other hand, we provide all knowledge aspects to construct the BHG and perform knowledge filtering automatically, which enables the aggregation nodes to retain all useful knowledge. The knowledge source $COMET_{2020}$ is proven most effective in enhancing emotional reasoning as it outperforms all other knowledge sources on all five datasets. Compared to $COMET_{2019}$, $COMET_{2020}$ trains a larger language model on a bigger social commonsense knowledge graph. Therefore, $COMET_{2020}$ is expected to generate more reliable knowledge for each utterance. These results show the importance of high-quality commonsense knowledge sources.

\subsubsection{BHG Variants}
We further compare the BHG-based methods in experimental settings such as "w/o future context" and "w/o emotions" to test model performance in scenarios where future contexts and emotion labels are unavailable. For ERC, the BHG's performance drops with only past context to a limited extent on most datasets. These results show that future dialogue can provide useful clues for the emotional reasoning of the current utterance in most cases. For CEE, the BHG performances on all knowledge resources drop less than 0.5\% without the emotional supervision signals, which is much less than the 1.7\% decrease on the previous state-of-the-art model KBCIN. We believe the appropriately infused commonsense knowledge from BHG can make up for the missing emotional information introduced by the labels. Finally, we compare the performance of MHGT and the vanilla HGT in modeling the BHG. During experiments for HGT, we unify the heterogeneous node dimensions by linearly projecting the high-dimension representations to low-dimensional spaces. According to the results, HGT significantly underperforms MHGT on all datasets. For example, we observe an over 3\% drop in Weighted-F1 for both $COMET_{2019}$ and $ConceptNet$ on IEMOCAP. MHGT retains the original semantic space for utterance representations, which preserves useful information to perform semantic-aware knowledge filtering and emotional reasoning, while HGT projects the representations to low-dimensional spaces and causes unnecessary loss of information.

\begin{table}[h]
\caption{Ablation studies for the BHG and other knowledge-based baseline models. "know." denotes the whole knowledge infusion architectures. "backward/forward aggr." denotes the backward/forward knowledge aggregation designs in BHG. We highlight top-2 performance drops in bold.}
\label{tab:ablation}
\begin{center}
\resizebox{.47\textwidth}{!}{
\begin{tabular}{l|cccc}
\toprule \textbf{Model} &  \textbf{IEMOCAP} &  \textbf{MELD} & \textbf{DailyDialog} & \textbf{EmoryNLP}\\ \midrule
CauAIN & 67.61 & 65.46 & 58.21 & --\\
\ w/o know. & 63.77 ($\downarrow$\textbf{3.84}) & 65.2 ($\downarrow$0.26) & 57.2 ($\downarrow$1.01) & --\\ \midrule
COSMIC & 65.28 & 65.21 & 58.48 & 38.11\\
\ w/o know. & 63.05 ($\downarrow$2.23) & 64.28 ($\downarrow$0.93) & 56.16 ($\downarrow$\textbf{2.32}) & 37.10 ($\downarrow$1.01)\\ \midrule
TODKAT & 61.33 & 65.47 & 54.62 & 38.69\\
\ w/o know. & 58.96 ($\downarrow$2.37) & 63.97 ($\downarrow$\textbf{1.5}) & 53.44 ($\downarrow$1.18) & 37.1 ($\downarrow$1.59)\\ \midrule
SKAIG & 66.96 & 65.18 & 59.75 & 38.88\\
\ w/o know. & 64.28 ($\downarrow$2.68) & -- & 58.86 ($\downarrow$0.89) & --\\ \midrule
BHG & 71.07 & 65.57 & 61.5 & 38.62\\
\ w/o backward aggr. & 68.54 ($\downarrow$2.53) & 64.81 ($\downarrow$0.76) & 60.18 ($\downarrow$1.32) & 38.07 ($\downarrow$0.55)\\
\ w/o forward aggr. & 68.07 ($\downarrow$3.0) & 64.41 ($\downarrow$1.16) & 59.3 ($\downarrow$2.3) & 36.95 ($\downarrow$\textbf{1.67})\\
\ w/o know. & 67.93 ($\downarrow$\textbf{3.14}) & 63.7 ($\downarrow$\textbf{1.87}) & 58.76 ($\downarrow$\textbf{2.74}) & 36.26 ($\downarrow$\textbf{2.36})\\
\bottomrule
\end{tabular}}
\end{center}
\end{table}

\subsection{Ablation Studies for Knowledge Infusion}
To further investigate the BHG's efficiency in knowledge infusion and the contributions of its components, we perform ablation studies on the BHG and other knowledge-based baseline methods, and the results are presented in Table \ref{tab:ablation}. To ensure fair comparisons, we select the BHG and four knowledge infusion methods that are all tested with the $COMET_{2019}$ knowledge source on the four ERC datasets.

\subsubsection{Knowledge Infusion Efficiency}
Firstly, we compare the performance drops between the models when the whole knowledge infusion architecture is removed, where more significant drops reflect the higher efficiency of the method in leveraging the same knowledge source. According to the results, "BHG w/o know." achieves top-2 performance drops on all datasets, and the drops exceed 2\% on three out of four datasets, while other knowledge infusion methods normally make substantial contributions to only one or two datasets. These results show that our unified BHG method can utilize the same knowledge source to enhance emotional reasoning on differently distributed data more efficiently than previous customized knowledge infusion methods.

\subsubsection{BHG Modules}
We further investigate the backward/forward knowledge infusion architectures in BHG by removing these modules and comparing the performance. According to the results, "BHG w/o backward aggr." and "BHG w/o forward aggr." both perform worse than BHG on all datasets. These results not only strengthen that infusing commonsense knowledge can enhance the modeling of inter-utterance relations, but further prove that a unified knowledge infusion module can be decoupled from the utterance modeling process and used for any conversation models in a plug-in manner. They also prove the necessity of splitting the knowledge filtering and interaction processes for previous and future contexts in the BHG architecture. In addition, "BHG w/o forward aggr." suffers from higher performance drops on all datasets than "BHG w/o backward aggr.", which shows that social commonsense knowledge is more useful in modeling inter-utterance relations for future contexts. These observations correspond with the widely recognized prior that the current utterance has more significant influences on future utterances along the dialogue flow~\citep{shen-etal-2021-directed}.

\begin{figure*}[htpb]
\centering
\includegraphics[width=14cm,height=7cm]{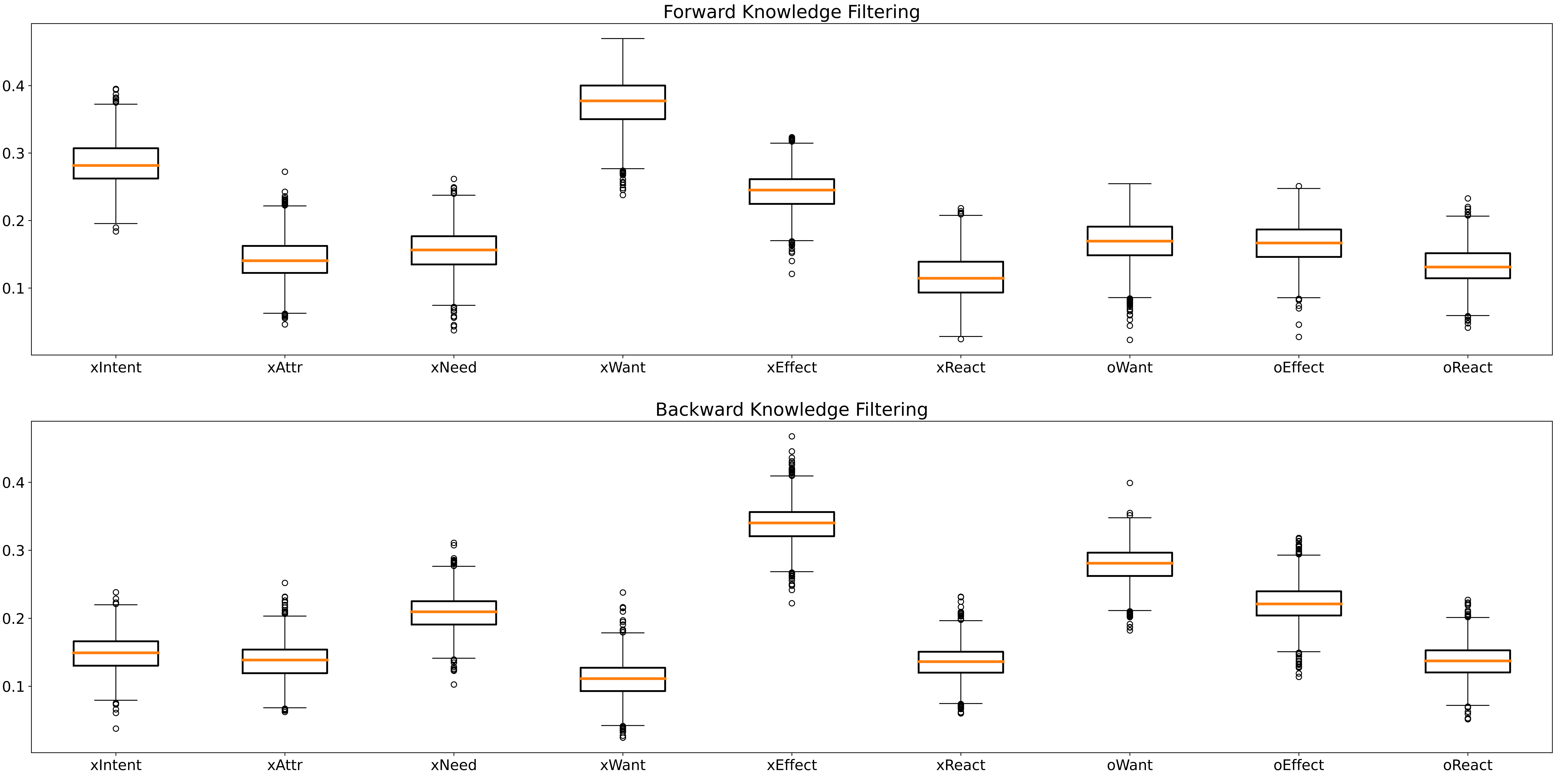}
\caption{Box plots of the knowledge filtering attention weights for nine knowledge aspects on the IEMOCAP test set. We use $COMET_{2019}$ as the knowledge source and MHGT as the BHG encoder. Orange lines denote the median numbers.}
\label{fig:box}
\end{figure*}

\subsection{Knowledge Filtering Analysis}
To provide an intuitive view of the knowledge filtering process, we record the dot-product attention weights between the knowledge nodes and knowledge aggregation nodes in the last MHGT layer for $COMET_{2019}$ on IEMOCAP test set and visualize their distributions in box plots. The results are presented in Figure \ref{fig:box}. Following the BHG structure, we split the attention weights from forward and backward knowledge filtering, and higher weight distributions reflect more contributions from the corresponding knowledge aspect.

\subsubsection{Overall Analysis}
For forward knowledge filtering, the most attended knowledge aspect \emph{xWant} has a median number over 0.3. \emph{xIntent} and \emph{xEffect} are also frequently utilized, with their median numbers over 0.2. These results show that forward knowledge interaction can benefit more from knowledge aspects reflecting the potential effect of the target utterance on the target speaker. On the other hand, backward knowledge filtering pays attention to the knowledge aspects reflecting the effect on both the target speaker and other participants. For example, \emph{xEffect} and \emph{oWant} are highly attended with median numbers over 0.3, and \emph{xNeed} and \emph{oEffect} also have median numbers of over 0.2. In comparisons of the box plots for forward and backward knowledge filtering, the BHG has very different patterns for selecting knowledge aspects. For example, \emph{xWant} knowledge is considered useful in forward knowledge filtering but receives much less attention in backward knowledge filtering. These observations further show the necessity of designing forward and backward knowledge aggregation nodes in BHG. In addition, some knowledge aspects such as \emph{xAttr} and \emph{oReact} have low attention distributions in both forward and backward knowledge filtering, possibly because these knowledge aspects can provide less relevant information for emotional reasoning.

\subsubsection{Comparison with Previous Methods}
Previous knowledge-based methods mostly perform knowledge filtering manually in the utilization of COMET$_{2019}$, where frequently utilized knowledge aspects include \emph{xWant}, \emph{xEffect}, \emph{xIntent}, \emph{oReact}, \emph{oWant}, and \emph{oEffect}~\citep{ghosal-etal-2020-cosmic,zhu-etal-2021-topic,li-etal-2021-past-present,DBLP:conf/ijcai/ZhaoZL22}. Most of these knowledge aspects are also proven effective in the automatic knowledge filtering of BHG, while some aspects such as \emph{xReact} and \emph{oReact} are less attended in both forward and backward knowledge filtering. These results indicate that some widely used knowledge aspects may not provide much useful information as expected. On the other hand, some previously ignored knowledge aspects are assigned higher attention scores in automatic knowledge filtering. For example, \emph{xNeed}, a less-used knowledge aspect in previous works, receives considerable attention in backward knowledge filtering. We expect the above analysis to guide the adjustment of future strategies for empirical methods.

\section{Related Work}
We discuss relevant background on conversational emotion reasoning, including conversation models and knowledge-based methods.
\subsection{Conversation Models}
Emotion reasoning in a conversation naturally requires modeling interactions between dialogue participants, known as intra- and inter-speaker dependencies~\citep{shen2021dialogxl}. Early works regarded dialogues as temporal flows and utilized RNNs to model the dialogue history and emotional dynamics of each dialogue participant~\citep{DBLP:conf/aaai/MajumderPHMGC19,ghosal-etal-2020-cosmic}. In other works, Transformer variants were also leveraged to model long-range dependencies in conversations~\citep{zhong-etal-2019-knowledge,zhang-etal-2020-knowledge}. Recent works mostly relied on PLMs such as RoBERTa~\citep{liu2019roberta} and XLNet~\citep{yang2019xlnet} to obtain utterance-level or conversation-level features~\citep{DBLP:conf/ijcai/BaoMWZH22,shen2021dialogxl}. Some other works~\citep{shen-etal-2021-directed,ishiwatari-etal-2020-relation,ghosal-etal-2019-dialoguegcn} modeled utterances as nodes and carefully designed graph structures on the dialogue to enable efficient message passing among utterances during the graph aggregation process. Other self-supervised architectures such as the Variational Autoencoder~\citep{10135132,ong2022discourse} were also used for modeling conversations or discourse information. 

\subsection{Commonsense Knowledge-based Methods}
Due to the implicitness of emotional expression in many scenarios, commonsense knowledge has been widely utilized to enhance emotional reasoning. One line of works \citep{zhong-etal-2019-knowledge,zhang-etal-2020-knowledge,xie-etal-2021-knowledge-interactive} extracted phrase-level concepts from the large-scale knowledge graph ConceptNet~\citep{speer2017conceptnet} to concatenate them with the token-level utterance representations, and Transformer-based models were used to perform utterance-knowledge interactions. More recent works leveraged the utterance-level knowledge from the generative knowledge source COMET~\citep{bosselut-etal-2019-comet,DBLP:conf/aaai/HwangBBDSBC21}. Some methods directly infused the knowledge into utterance representations~\citep{ghosal-etal-2020-cosmic,zhu-etal-2021-topic}, followed by neural network-based knowledge infusion modules. Other works utilized the knowledge as edge representations in the dialogue graphs to model inter-utterance relations between utterance nodes~\citep{li-etal-2021-past-present,zhao2022knowledge,DBLP:conf/ijcai/LiM0L0C0Z22} or trace emotion casual clues~\citep{DBLP:conf/ijcai/ZhaoZL22}, which obtained outstanding performance in both emotion detection~\citep{li-etal-2021-past-present} and emotion casual detection tasks~\citep{zhao2022knowledge,DBLP:conf/ijcai/LiM0L0C0Z22}. In addition, the pre-trained knowledge adapters were also used for knowledge infusion into PLM-based conversation models~\citep{yang2023cluster}.

\section{Conclusion and Future Work}
This paper proposes a bipartite heterogeneous graph for enhancing emotional reasoning with commonsense knowledge. We model the utterance representations and knowledge representations as heterogeneous nodes and design a BHG for commonsense knowledge infusion. In addition, we propose a multi-dimensional heterogeneous graph Transformer to perform graph reasoning to retain unchanged feature spaces for heterogeneous node types. Experiments show that BHG-based methods outperform state-of-the-art knowledge infusion methods on five datasets across two conversational emotional reasoning tasks. The BHG also shows generalized knowledge infusion ability with higher efficiency. Further analysis proves that previous empirical knowledge filtering methods do not guarantee to provide the most useful knowledge information.

In future work, we will test our BHG-based method on more conversation models, such as graph-based models, to further examine their generalizability. We will also explore simultaneous knowledge infusion from multiple knowledge sources in a unified BHG framework, which enables the model to reason on several knowledge types to enhance its performance. 

\begin{acks}
This work is supported in part by the New Energy and Industrial Technology Development Organization (NEDO) project under grant JPNP20006 and the University of Manchester President’s Doctoral Scholar award.
\end{acks}

\bibliographystyle{ACM-Reference-Format}
\bibliography{anthology,sample-base}


\end{document}